\documentclass{article}     
\usepackage{microtype}
\usepackage{graphicx}
\usepackage{subcaption}
\usepackage{eqparbox}
\usepackage{booktabs} 
\usepackage{hyperref}
\usepackage{soul}

\usepackage[accepted]{icml2024}
\usepackage{amsmath}
\usepackage{amssymb}
\usepackage{mathtools}
\usepackage{amsthm}
\usepackage[capitalize,noabbrev]{cleveref}
\theoremstyle{plain}

\theoremstyle{definition}

\theoremstyle{remark}

\usepackage[textsize=tiny]{todonotes}
\newcommand\shrparams{\ensuremath{\Psi}}
\newcommand\taskparams{\ensuremath{\Phi_t}}
\newcommand\gateparams{\ensuremath{\alpha}}
\newcommand\shrfeat{\ensuremath{\psi}}
\newcommand\taskfeat{\ensuremath{\varphi_t}}
\newcommand\combwgt{\ensuremath{\beta_t}}
\newcommand{\methodname}{InterroGate}
\usepackage{nicematrix}
\icmltitlerunning{\methodname: Learning to Share, Specialize, and Prune Representations for Multi-task Learning}
\begin{document}
\def\@fnsymbol#1{\ensuremath{\ifcase#1\or \dagger\or \ddagger\or
\mathsection\or \mathparagraph\or \|\or **\or \dagger\dagger
\or \ddagger\ddagger \else\@ctrerr\fi}}
\twocolumn[
\icmltitle{\methodname: Learning to Share, Specialize, and Prune \texorpdfstring{Representations \\}{Representations} for Multi-task Learning}
\icmlsetsymbol{equal}{*}
\begin{icmlauthorlist}
\icmlauthor{Babak Ehteshami Bejnordi}{yyy}
\icmlauthor{Gaurav Kumar}{comp}
\icmlauthor{Amelie Royer}{yyy}
\icmlauthor{Christos Louizos}{yyy} \\
\icmlauthor{Tijmen Blankevoort \texorpdfstring{$^{\dagger}$}{d}}{yyy}
\icmlauthor{Mohsen Ghafoorian}{BBB}
\end{icmlauthorlist}
\icmlaffiliation{yyy}{Qualcomm AI Research\texorpdfstring{$^\text{*}$}{*}, Amsterdam, The Netherlands.}
\icmlaffiliation{comp}{Qualcomm AI Research, Hyderabad, India.}
\icmlaffiliation{BBB}{XR Labs, Qualcomm Technologies Inc., Amsterdam, The Netherlands}
\icmlkeywords{Machine Learning, ICML}
\vskip 0.3in
]
\printAffiliationsAndNotice{}  
\begin{abstract}
Jointly learning multiple tasks with a unified model can improve accuracy and data efficiency, but it faces the challenge of task interference, where optimizing one task objective may inadvertently compromise the performance of another.
A solution to mitigate this issue is to allocate task-specific parameters, free from interference, on top of shared features. However, manually designing such architectures is cumbersome, as practitioners need to balance between the overall performance across all tasks and the higher computational cost induced by the newly added parameters.
In this work, we propose \textbf{\methodname}, a novel MTL architecture designed to mitigate task interference while optimizing inference computational efficiency.
We employ a learnable gating mechanism to automatically balance the shared and task-specific representations while preserving the performance of all tasks.
Crucially, the patterns of parameter sharing and specialization dynamically learned during training, become fixed at inference, resulting in a static, optimized MTL architecture.
Through extensive empirical evaluations, we demonstrate SoTA results on three MTL benchmarks using convolutional as well as transformer-based backbones on CelebA, NYUD-v2, and PASCAL-Context.
\end{abstract}
\setlength{\belowcaptionskip}{-10pt}
\setlength{\abovecaptionskip}{-10pt}
\setlength{\floatsep}{-0pt}
\section{Introduction}
\label{sec:intro}
Multi-task learning (MTL) involves learning multiple tasks concurrently with a unified architecture.
By leveraging the shared information among related tasks, MTL has the potential to improve accuracy and data efficiency.
In addition, learning a joint representation reduces the computational and memory costs of the model at inference as visual features relevant to all tasks are extracted only once:
This is crucial for many real-life applications where a single  device is expected to solve multiple tasks simultaneously under strict compute constraints (e.g. mobile phones, extended reality, self-driving cars, etc.).
Despite these potential benefits, in practice, MTL is often met with a key challenge known as \textit{negative transfer} or \textit{task interference}~\cite{zhao2018modulation}, which refers to the phenomenon where the learning of one task negatively impacts the learning of another task during joint training.
While characterizing and solving task interference is an open issue~\cite{Wang2018CharacterizingAA,royer2023scalarization}, there exist two major lines of work to mitigate this problem: \textbf{(i)} Multi-task Optimization (MTO) techniques aim to balance the training process of each task, while \textbf{(ii)} architectural designs carefully allocate shared and task-specific parameters to reduce interference.

Specifically, MTO approaches balance the losses/gradients of each task to mitigate the extent of gradient conflicts while optimizing the shared features.
However, the results may still be compromised when the tasks rely on inherently different visual cues, making sharing parameters difficult.
For instance, text detection and face recognition require learning very different texture information and object scales.
An alternative and orthogonal research direction is to allocate additional task-specific parameters, on top of shared generic features, to bypass task interference.
In particular, several state-of-the-art methods have proposed task-dependent selection and adaptation of shared features ~\cite{Guo2020LearningTB, sun2020adashare,Wallingford2022TaskAP, rahimian2023dynashare}.  
However, the dynamic allocation of task-specific features is usually performed one task at a time, and solving all tasks still requires multiple forward passes. 
Alternatively, Mixture of Experts (MoE) have also been employed to reduce the computational cost of MTL by dynamically routing inputs to a subset of experts \cite{ma2018modeling,hazimeh2021dselect, fan2022m3vit, chen2023mod}.
However, the input-dependent routing of MoE is typically hard to efficiently deploy at inference, particularly with batched execution \cite{sarkar2023edge, yi2023edgemoe}.

In contrast to previous dynamic architectures, we learn to balance shared and task-specific features jointly for all tasks, which allows us to predict all task outputs in a single forward pass. 
In addition, we propose to regulate the expected inference computational cost through a budget-aware regularization during training.
By doing so, we aim to depart from a common trend in MTL that heavily focuses on accuracy while neglecting computational efficiency \cite{misra2016cross, vandenhende2020mti}.

In this paper, we introduce \methodname, a novel MTL architecture
to mitigate task interference while optimizing computational efficiency during inference. Our method learns the per-layer optimal balance between sharing and specializing representations for a desired computational budget.
In particular, we leverage a shared network branch which is used as a general communication channel through which the task-specific branches interact with each other.
This communication is enabled through a novel gating mechanism which learns for each task and layer to select parameters from either the shared branch or their task-specific branch. To enhance the learning of the gating behaviour, we harness single task baseline weights to initialize task-specific branches.

\methodname{} primarily aims to optimize efficiency in the inference phase, crucial in real-world applications. While the gate dynamically learns to select between a large pool of task-specific and shared parameters during training, at inference, the learned gating patterns are static and thus can be used to prune the unselected parameters in the shared and task-specific branches: As a result, \methodname{} collapses to a simpler, highly efficient, static architecture at inference time, suitable for batch processing.
We control the trade-off between inference computational cost and multi-task performance, by regularizing the gates using a sparsity objective.
In summary, our contributions are as follows:

\begin{itemize}
\item We propose a novel multi-task learning framework that learns the optimal parameter sharing and specialization patterns for all tasks, in tandem with the model parameters, enhancing multi-task learning efficiency and effectiveness.

\item We enable a training mechanism to control the trade-off between multi-task performance and inference compute cost. Our proposed gating mechanism finds the optimal balance between selecting shared and specialized parameters for each task, within a desired computational
budget, controlled with a sparsity objective. This, subsequently, enables a simple process for creating a range of models on the efficiency/accuracy trade-off spectrum, as opposed to most other MTL methods.

\item Our proposed method is designed to optimize inference-time efficiency. Post-training, the unselected parameters by the gates are pruned from the model, resulting in a simpler, highly efficient neural network. In addition, our feature fusion strategy allows to predict all tasks in a single forward pass, critical in many real-world applications.
\item Through extensive empirical evaluations, we report SoTA results consistently on three multi-tasking benchmarks with various convolutional and transformer-based backbones. We then further investigate the proposed framework through ablation experiments.
\end{itemize}

\section{Related Work}
\label{sec:related}
    \textbf{Multi-task Optimization} (MTO) methods aim to automatically balance the different tasks when optimizing shared parameters to maximize average performance.
    Loss-based methods~\cite{kendall2018multi,Liu2022AutoLambdaDD} are usually scalable and adaptively scale task losses based on certain statistics (e.g. task output variance); Gradient-based methods~\cite{sener2018multi,chen2018gradnorm,Chen2020JustPA,liu2021conflict,Javaloy2022RotoGradGH} are more costly in practice as they require storing a gradient per task, but usually yield higher performance.
    Orthogonal to these optimization methods, several research directions investigate how to design architectures that inherently mitigate task-interference, as described below.

    \textbf{Task Grouping}  approaches investigate which groups of tasks can safely share encoder parameters without task interference.
    For instance \cite{taskgrouping, Fifty2021EfficientlyIT} identify ``task affinities" as a guide to isolate parameters of tasks most likely to interfere with one another. Similarly, \cite{Guo2020LearningTB} apply neural architecture search techniques to design MTL architectures.
    However, exploring these large architecture search spaces is a costly process.

    \textbf{Hard Parameter Sharing} works such as Cross-Stitch~\cite{misra2016cross}, MTAN~\cite{liu2019end}  or MuIT~\cite{Bhattacharjee2022MuITAE} propose to learn the task parameter sharing design alongside the model features.
    However, most of these works mainly focus on improving the accuracy of the model while neglecting the computational cost:
    For instance, \cite{misra2016cross, Gao2020MTLNASTN} require a full network per task and improve MTL performance through lightweight adapters across task branches.
    \cite{liu2019end,Bhattacharjee2022MuITAE} use task-specific attention module on top of a shared feature encoder, but the cost of the task-specific decoder heads often dominates the final architecture.
    \begin{figure*}[htbp!]
\centering
\includegraphics[width=0.95\linewidth]{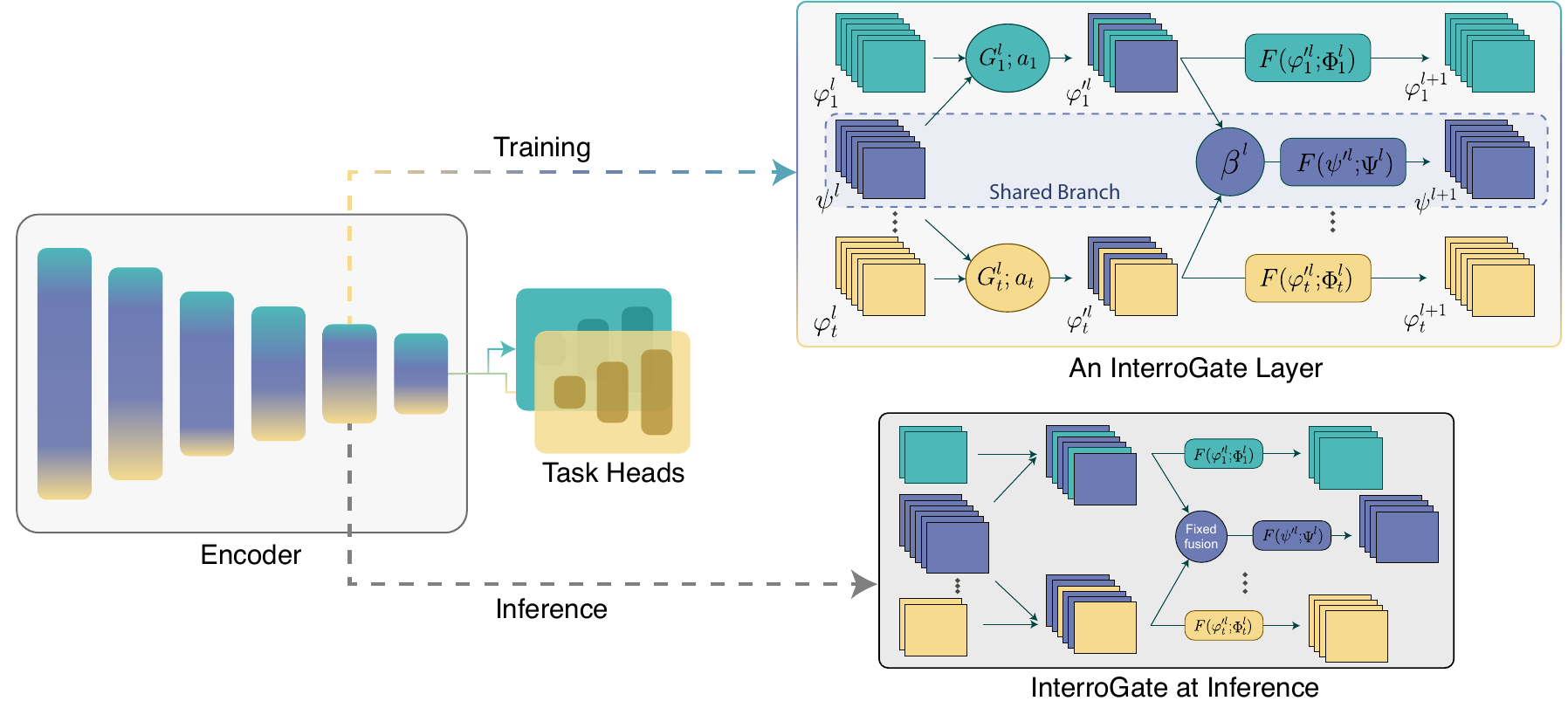}

\caption{\textbf{Overview of the proposed \methodname{} framework}: The original encoder layers are substituted with \methodname{} layers. The input to the layer is $t+1$ feature maps, one shared representation and $t$ task-specific representations. To decide between shared $\shrfeat^\ell$ or task-specific $\taskfeat^\ell$ features, each task relies on its own gating module $G_t^\ell$. The resulting channel-mixed feature-map $\taskfeat^{\prime \ell}$ is then fed to the next task-specific layer. The input to the shared branch for the next layer is constructed by linearly combining the task-specific features of all tasks using the learned parameter $\beta_t^{\ell}$. During inference, the parameters (shared or task-specific) that are not chosen by the gates are removed from the model, resulting in a plain neural network architecture.}
\label{fig:overview}
\vskip -0.1in
\end{figure*}

    \textbf{Conditional Compute}  approaches learn task-specific gating of model parameters:
    For instance \cite{sun2020adashare, Wallingford2022TaskAP} learn to select a subset of the most relevant layers  when adapting the network to a new downstream task.
    Piggyback \cite{mallya2018piggyback} adapts a pretrained network to multiple tasks by learning a set of per-task sparse masks for the network parameters. Similarly, \cite{Berriel2019BudgetAwareAF} select the most relevant feature channels using learnable gates.
    Finally, Mixture of Experts (MoE)~\cite{ma2018modeling,hazimeh2021dselect, fan2022m3vit, chen2023mod}
    leverage sparse gating to select a subset of experts for each input example.

    Nevertheless, due to the dynamic nature of these works at inference, obtaining all task predictions is computationally inefficient as it often requires one forward pass through the model per task. Therefore, these methods are less suited for MTL settings requiring solving all tasks concurrently. Additionally, dynamic expert selection in MoEs requires either storing all expert weights on-chip, increasing memory demands, or frequent off-chip data transfers to load necessary experts, leading to significant overheads, complicating their efficient inference on resource-constrained devices \cite{sarkar2023edge, yi2023edgemoe}.
    In contrast, \methodname{} focuses on optimizing efficiency and is capable of addressing all tasks simultaneously, aligning with many practical real-world needs. \methodname{} employs learned gating patterns to prune unselected parameters, resulting in a more streamlined and efficient static architecture during inference, well-suited for batch processing.
    Finally, closest to our work is \cite{Bragman2019StochasticFG}, which proposes a probabilistic allocation of convolutional filters as task-specific or shared. However, this design only allows for the shared features to send information to the task-specific branches.
    In contrast, our gating mechanism allows for information to flow in any direction between the shared and task-specific features, thereby enabling cross-task transfer in every layer.

\section{\methodname}
\label{sec:method}
Given $T$ tasks, we aim to learn a flexible allocation of shared and task-specific parameters, while optimizing the trade-off between accuracy and efficiency.
Specifically, an \methodname{} model is characterized by task-specific parameters $\{\taskparams\}_{t=1}^{T}$ and shared parameters $\shrparams$. In addition, discrete gates (with parameters $\gateparams$) are trained to only select a subset of the most relevant features in both the shared and task-specific branches, thereby reducing the model's computational cost.
Under this formalism, the model and gate parameters are trained end-to-end by minimizing the classical MTL objective:
\vspace{-2pt}
\begin{align}
\label{eq:accuracy}
\mathcal{L}(\{\taskparams\}_{t=1}^{T}, \shrparams, \gateparams) = \sum_{t=1}^{T} \omega_t\  \mathcal{L}_t(X, Y_t; \taskparams, \shrparams, \gateparams),
\end{align}
where 
\( X \) and \( Y_t \) are the input data and corresponding labels for task \( t \), $\mathcal{L}_t$ represents the loss function associated to task $t$, and \( \omega_t \) are hyperparameter coefficients which allow for balancing the importance of each task in the overall objective.
In the rest of the section, we describe how we learn and implement the feature-level routing mechanism characterized by $\gateparams$. We focus on convolutional architectures in \hyperref[sec:methodcnn]{Section \ref{sec:methodcnn}}, and discuss the case of transformer-based models in \hyperref[methodvit]{Appendix \ref{methodvit}}.

\subsection{Learning to Share, Specialize and Prune}
\label{sec:methodcnn}
\hyperref[fig:overview]{Figure \ref{fig:overview}} presents an overview of the proposed \methodname{} framework. 
Formally, let $\shrfeat^\ell \in \mathbb{R}^{C^\ell \times W^\ell \times H^\ell}$ and $\taskfeat^\ell \in \mathbb{R}^{C^\ell \times W^\ell \times H^\ell}$ represent the shared and task-specific features at layer $\ell$ of our multi-task network, respectively.
In each layer $\ell$, the gating module $G_t^\ell$ of task $t$ selects relevant channels from either $\shrfeat^{\ell}$ and $\taskfeat^\ell$. The output of this hard routing operation yields features $\taskfeat^{\prime \ell}$:
\begin{align}
\label{eq:taskmix}
\varphi^{\prime \ell}_t = G_t^\ell (\alpha_t^\ell) \odot \taskfeat^\ell + (1 - G_t^\ell (\alpha_t^\ell)) \odot \shrfeat^\ell,
\end{align}
where $\odot$ is the Hadamard product and $ \alpha_t^\ell \in {\mathbb R}^{C^\ell}$ denotes the learnable gate parameters for task $t$ at layer $\ell$ and the gating module $G_t^\ell$ outputs a vector in $\{0, 1\}^{C^\ell}$, encoding the binary selection for each channel. These intermediate features are then fed to the next task-specific layer to form the features $\taskfeat^{\ell+1} = F(\taskfeat^{\prime \ell}; \taskparams^\ell)$.

Similarly, we construct the shared features of the next layer $\ell + 1$ by mixing the previous layer's task-specific feature maps. However, how to best combine these $T$ feature maps is a harder problem than the pairwise selection described in \hyperref[eq:taskmix]{(\ref{eq:taskmix})}.
Therefore, we let the model learn its own soft combination weights and the resulting mixing operation for the shared features is defined as follows:
\vspace{-2pt}
\begin{equation}
\label{shrmix}
\shrfeat^{\prime \ell} = \sum_{t=1}^{T} \underset{t=1 \dots T}{\text{softmax}}(\combwgt^\ell) \odot \taskfeat^{\prime \ell},
\end{equation}
where $\beta^\ell \in \mathbb{R}^{C^\ell \times T}$ denotes the learnable parameters used to linearly combine the task-specific features and form the shared feature map of the next layer.
Similar to the task-specific branch, these intermediate features are then fed to a convolutional block to form the features $\shrfeat^{\ell+1} = F(\shrfeat^{\prime \ell}; \shrparams^\ell)$.
Finally, note that there is no direct information flow between the shared features of one layer to the next, i.e., $\shrfeat^\ell$ and $\shrfeat^{\ell + 1}$: Intuitively, the shared feature branch can be interpreted as a general communication channel which the task-specific branches communicate with one another.

\subsection{Implementing the Discrete Routing Mechanism}

During \textbf{training}, the model features and gates are trained jointly and end-to-end. 
In \hyperref[eq:taskmix]{(\ref{eq:taskmix})}, the gating modules $G_t^\ell$ each output a binary vector over channels in $\{0, 1\}^C$, where 0 means choosing the shared feature at this channel index, while 1 means choosing the specialized feature for the respective task $t$.
In practice, we implement $G$ as a sigmoid operation applied to the learnable parameter $\alpha$, followed by a thresholding operation at 0.5.
Due to the non-differentiable nature of this operation, we adopt the straight-through estimation (STE) during training~\cite{bengio2013estimating}:  In the backward pass, STE approximates the gradient flowing through the thresholding operation as the identity function.

At \textbf{inference}, since the gate modules do not depend on the input data, our proposed \methodname{} method converts to a static neural network architecture, where feature maps are pruned following the learned gating patterns:
To be more specific, for a given layer $\ell$ and task $t$, we first collect all channels for which the gate $G_t^\ell(\alpha_t^\ell)$ outputs 0; Then, we simply prune the corresponding task-specific weights in $\taskparams^{\ell-1}$.
Similarly, we can prune away weights from the shared branch $\shrparams^{\ell-1}$ if the corresponding channels are never chosen by any of the tasks in the mixing operation of \hyperref[eq:taskmix]{(\ref{eq:taskmix})}. The pseudo-code for the complete unified encoder forward-pass is detailed in \hyperref[alg:cap]{Appendix \ref{alg:cap}}.

\subsection{Sparsity Regularization}
During training, we additionally control the proportion of shared versus task-specific features usage by regularizing the gating module $G$.
This allows us to reduce the computational cost, as more of the task-specific weights can be pruned away at inference.
We implement the regularizer term as a hinge loss over the gating activations for task-specific features:
\vspace{-2pt}
\begin{equation}
\label{eq:sparsity}
\mathcal{L}_{\text{sparsity}}(\alpha ) = \frac{1}{T}\sum_{t=1}^{T} \text{max}\left(0,\ \frac{1}{L} \sum_{\ell=1}^{L} \sigma(\gateparams_t^\ell)-\tau_t\right),
\end{equation}
\vspace{-2pt}
where $\sigma$ is the sigmoid function and $\tau_t$ is a task-specific hinge target. The parameter $\tau$ allows to control the proportion of active gates at each specific layer by setting a soft upper limit for active task-specific parameters. A lower hinge target value encourages more sharing of features while a higher value gives more flexibility to select task-specific features albeit at the cost of higher computational cost.

Our final training objective is a combination of the multi-task objective and sparsity regularizer:
\vspace{-2pt}
\begin{align}
\mathcal{L} = \mathcal{L}(\{\taskparams\}_{t=1}^{T},\Psi,\alpha,\beta) + \lambda_{\text{s}} \mathcal{L}_{\text{sparsity}}(\alpha ),
\end{align}
where $\lambda_s$ is a hyperparameter balancing the two losses. \\

\section{Experiments}
\label{sec:exp}
\subsection{Experimental Setup}

\paragraph{Datasets and Backbones.}
We evaluate the performance of \methodname{} on three popular datasets: CelebA \cite{celeba}, NYUD-v2 \cite{silberman2012indoor}, and PASCAL-Context \cite{chen2014detect}.  CelebA is a large-scale face attributes dataset, consisting of more than 200k celebrity images, each labeled with 40 attribute annotations. We consider the age, gender, and clothes attributes to form three output classification tasks for our MTL setup and use binary cross-entropy to train the model. The NYUD-v2 dataset is designed for semantic segmentation and depth estimation tasks. It comprises 795 train and 654 test images taken from various indoor scenes in New York City, with pixel-wise annotation for semantic segmentation and depth estimation. Following recent work \cite{xu2018pad, zhang2019pattern, maninis2019attentive}, we also incorporate the surface normal prediction task, obtaining annotations directly from the depth ground truth. We use the mean
intersection over union (mIoU) and root mean square error (rmse) to evaluate the semantic segmentation and depth estimation tasks, respectively. We use the mean error (mErr) in the predicted angles to evaluate the surface normals.
Finally, the PASCAL-Context dataset is an extension of the PASCAL VOC 2010 dataset \cite{everingham2010pascal} and provides a comprehensive scene understanding benchmark by labeling images for semantic segmentation, human parts segmentation, semantic edge detection, surface normal estimation, and saliency detection. The dataset consists of 4,998 train images and 5,105 test images.
The semantic segmentation, saliency estimation, and human part segmentation tasks are evaluated using mIoU. Similar to NYUD, mErr is used to evaluate the surface normal predictions. 

We use ResNet-20 \cite{he2016deep} 
as the backbone in our CelebA experiments, with simple linear heads for the task-specific predictions.
For the NYUD-v2 dataset, we use ResNet-50 with dilated convolutions and HRNet-18  following \cite{vandenhende2021multi}. We also present results using a dense prediction transformer (DPT) \cite{ranftl2021vision}, with a ViT-base and -small backbone.
Finally, on PASCAL-Context, we use a ResNet-18 backbone.
For both NYUD and PASCAL, we use dense prediction heads to output the task predictions, as described in \hyperref[Implementation]{Appendix \ref{Implementation}}.

\paragraph{SoTA Baselines and Metrics.}

To establish upper and lower bounds of MTL performance, we always compare our models to the Single-Task baseline (STL), which is the performance obtained when training an independent network for each task, as well as the uniform MTL baseline where the model's encoder (backbone) is shared by all tasks.

In addition, we compare \methodname~to encoder-based methods including Cross-stitch \cite{misra2016cross} and MTAN \cite{liu2019end}, as well as  
MTO approaches such as uncertainty weighting \cite{kendall2018multi}, DWA \cite{liu2019end}, and  Auto-\(\lambda\) \cite{Liu2022AutoLambdaDD}, PCGrad \cite{Yu2020GradientSF}, CAGrad \cite{liu2021conflict}, MGDA-UB \cite{sener2018multi}, and RLW \cite{Lin2022ReasonableEO}.
Following \cite{maninis2019attentive}, our main metric is the multi-task performance \(\Delta_\text{MTL}\) of a model \(m\) as the averaged normalized drop in performance w.r.t. the single-task baselines \(b\):
\vspace{-2pt}
\begin{equation}
\Delta_\text{MTL}=\frac{1}{T} \sum_{i=1}^{T}(-1)^{l_{i}}\left(M_{m, i}-M_{b, i}\right) / M_{b, i}
\end{equation}
where \(l_{i}=1\) if a lower value means better performance
for metric \(M_{i}\) of task \(i\), and 0 otherwise. Furthermore, similar to \cite{navon2022multi}, we compute the mean rank (MR) as the average rank of each method across the different tasks, where a lower MR indicates better performance.  All reported results for \methodname{} and baselines are averaged across 3 random seeds.

\begin{figure*}[!htbp]
\centering
\includegraphics[width=1\linewidth]{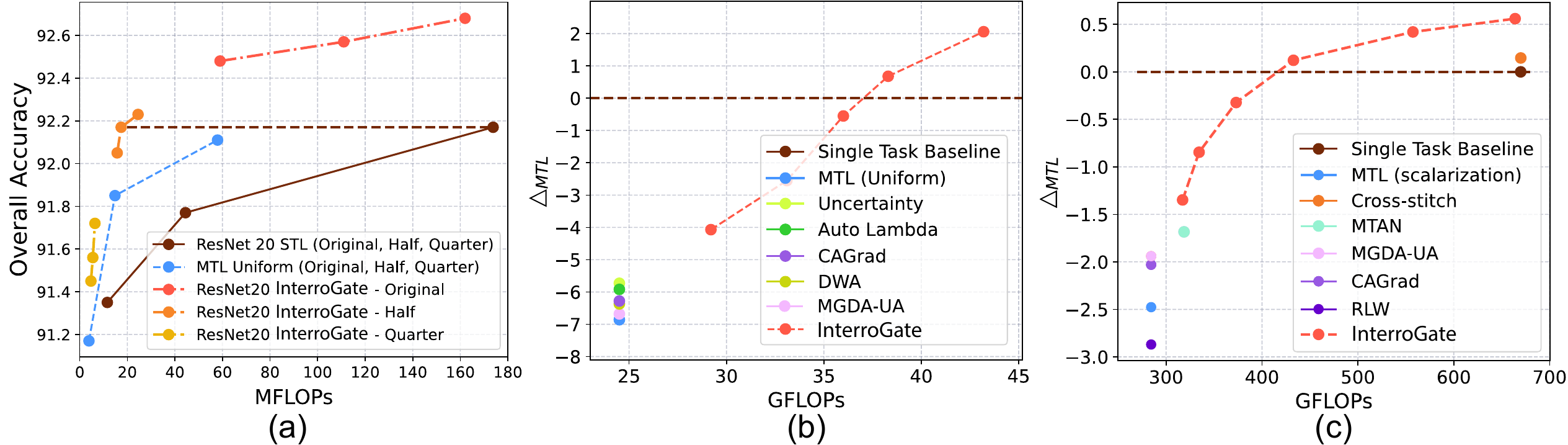}
\caption{Accuracy vs. floating-point operations (FLOP) trade-off curves for \methodname{} and SoTA MTL methods. (a) Results on CelebA using ResNet-20 backbone at three different widths (Original, Half, and Quarter). (b) NYUD-v2 using HRNet-18 backbone, and (c) ResNet-18 on PASCAL-Context. To avoid clutter, we present the six highest-performing MTL baselines in \textit{(b)} \& \textit{(c)}. The single task baseline in \textit{(b)} has 65.1 GFLOPs.}
\label{fig:all_curves}
\vskip -0.1in
\end{figure*}

Finally, to generate the trade-off curve between MTL performance and compute cost of \methodname{} in \hyperref[fig:all_curves]{Figure~\ref{fig:all_curves}} and all the tables, we sweep over the gate sparsity regularizer weight, $\lambda_s$, in the range of $\{1,3,5,7,10\}\cdot10^{-2}$.
The task-specific targets $\tau$ in \hyperref[eq:sparsity]{(\ref{eq:sparsity})} also impact the computation cost.  
Intuitively, tasks with significant performance degradation benefit from more task-specific parameters, i.e., from higher values of the hyperparameters  $\tau_t$. 
We analyze the gating patterns for sharing and specialization in section~\ref{gatingpattern} and, through ablation experiments, we further discuss the impact of sparsity targets $\{\tau_t\}_{t=1}^T$ in \hyperref[app:sparsity]{Appendix \ref{app:sparsity}}.

While \methodname{} primarily aims at improving the inference cost efficiency, we also measure and report training time comparison between our method and the baselines, on the PASCAL-Context \cite{chen2014detect} dataset in \hyperref[app:timing]{Appendix~\ref{app:timing}}.

\paragraph{Training Pipeline.}

For initialization, we use pretrained ImageNet weights for the single-task and multi-task baseline.
For \methodname, the shared branch is initialized with ImageNet weights while the task-specific branches are with their corresponding single-task weights.
Finally, we discover that employing a separate optimizer for the gates improves the convergence of the model. 
We further describe training hyperparameters in \hyperref[Implementation]{Appendix \ref{Implementation}}. All experiments were conducted on a single NVIDIA V100 GPU and we use the same training setup and hyperparameters across all MTL approaches included in our comparison.

\subsection{Results}
In this section, we present the results of \methodname, single-task (STL), multi-task baselines, and the competing MTL approaches, on the CelebA, NYUD-v2, and PASCAL-Context datasets. In Section \ref{ablations}, we will also present several ablation studies on \textbf{(i)} how we design the sparsity regularization loss, \textbf{(ii)} on the qualitative sharing/specialization patterns learned by the gate, and finally, \textbf{(ii)} on the impact of the backbone model capacity on MTL performance. As mentioned earlier, we also present an ablation on the impact of the sparsity targets $\{\tau_t\}_{t=1}^T$ in \hyperref[app:sparsity]{Appendix \ref{app:sparsity}}.

\subsubsection{CelebA}

\hyperref[fig:all_curves]{Figure \ref{fig:all_curves}a} shows the trade-off between MTL performance and the computational cost (FLOPs) for \methodname, MTL uniform, and STL baselines on the CelebA dataset, for 3 different widths for the ResNet-20 backbone: quarter, half, and original capacity.
We report the detailed results in Table \ref{table:full_celebA},  \hyperref[app:results]{Appendix~\ref{app:results}}. 
\methodname{} 
outperforms MTL uniform and STL baselines with higher overall accuracy at a much lower computational cost.
Most notably, the performance of \methodname{} with ResNet-20 half width at only 14.8 MFlops matches the performance of STL with 174 MFlops.
Finally, we further discuss the behavior of \methodname{} and MTL baselines across different model capacities in the ablation experiment in section~\ref{cap_ablation}.

\vspace{-2mm}
\subsubsection{NYUD-v2}
\hyperref[table:nyud1]{Table \ref{table:nyud1}} and \ref{table:nyud2} present the results on the NYUD-v2 dataset, using the HRNet-18 and ResNet-50 backbones, respectively. \hyperref[fig:all_curves]{Figure \ref{fig:all_curves}b} illustrates the trade-off between $\Delta_\text{MTL}$ and computational cost of various methods using the HRNet-18 backbone. We additionally report the parameter count and the standard deviation of the $\Delta_\text{MTL}$ scores in \hyperref[table:nyud1a]{Tables~\ref{table:nyud1a}} and \ref{table:nyud2a}, in \hyperref[app:results]{Appendix \ref{app:results}}. As can be seen, most MTL methods improve the accuracy on the segmentation and depth estimation tasks, while surface normal prediction significantly drops. 
While MTL uniform and MTO strategies operate at the lowest computational cost by sharing the full backbone, they fail to compensate for this drop in performance. MTO approaches often show mixed results across the two backbones. While CAGrad \cite{liu2021conflict} and uncertainty weighting \cite{kendall2018multi} are the best performing MTO baselines using ResNet-50 and HRNet-18 backbones, respectively, they show negligible improvement over uniform MTL when applied to the alternate backbones.
In contrast, among the encoder-based methods, Cross-stitch largely retains performance on normal estimation 
and achieves a positive $\Delta_\text{MTL}$ score of  +1.66. However, this comes at a substantial computational cost and parameter count, close to that of the STL baseline.
In comparison, \methodname{} achieves an overall $\Delta_\text{MTL}$ score of +2.06 and +2.04 using HRNet-18 and ResNet-50, respectively, at a lower computational cost. At an equal parameter count of 92.4 M, \methodname{} surpasses MTAN using the ResNet-50 backbone, exhibiting a $\Delta_\text{MTL}$ score of +1.16, in contrast to MTAN's -0.84.
\setlength{\tabcolsep}{+0.5 pt}
\begin{table}[!htb]
\caption{Performance comparison on NYUD-v2 using HRNet-18.}
\label{table:nyud1}
\vskip -0.2in
\begin{center}
\begin{small}
\begin{sc}
\tiny
\begin{tabular*}{\columnwidth}{@{\extracolsep{\fill}}lccccccc}
\toprule
Model & Semseg $\uparrow$ & Depth $\downarrow$ & Normals $\downarrow$ & $\Delta_\text{MTL}$ (\%) $\uparrow$ & Flops (G) & MR$\downarrow$ \\
\midrule
STL & 41.70 & 0.582 & \textbf{18.89} & 0 & 65.1 & 8.0 \\
MTL (Uni.) & 41.83 & 0.582 & 22.84 & -6.86  & 24.5 & 11.0 \\
DWA & 41.86 & 0.580 & 22.61 &  -6.29  & 24.5 & 8.7 \\
Uncertainty & 41.49 & 0.575 & 22.27 & -5.73  & 24.5 & 8.3 \\
Auto-$\lambda$ & 42.71 & 0.577 & 22.87 &  -5.92  & 24.5 & 8.0 \\
RLW & 42.10 & 0.593 & 23.29 &  -8.09 & 24.5 & 11.7 \\
\midrule
PCGrad & 41.75 & 0.581 & 22.73 &  -6.70 & 24.5 & 10.3 \\
CAGrad & 42.31 & 0.580 & 22.79 &  -6.28 & 24.5 & 8.7 \\
MGDA-UB & 41.23 & 0.625 & 21.07 &  -6.68 & 24.5 & 11.3 \\

\midrule
\methodname & \textbf{43.58} & \textbf{0.559} & 19.32 & \textbf{+2.06} & 43.2 & \textbf{1.3} \\
\methodname & 42.95 & 0.562 & 19.73 & +0.68  & 38.3 & 2.3 \\
\methodname & 42.36 & 0.564 & 20.04 & -0.55 & 36.0 & 4.0 \\
\methodname & 42.73 & 0.575 & 21.01 & -2.55 & 33.1 & 4.0 \\
\methodname & 42.35 & 0.575 & 21.70 & -4.07 & 29.2  & 5.7 \\
\bottomrule
\end{tabular*}
\end{sc}
\end{small}
\end{center}
\end{table}

\begin{table}[!htb]
\caption{Performance comparison on NYUD-v2 using ResNet-50.}
\label{table:nyud2}
\vskip 0.in
\begin{center}
\begin{small}
\begin{sc}
\tiny
\begin{tabular*}{\columnwidth}{@{\extracolsep{\fill}}lccccccc}
\toprule
Model & Semseg $\uparrow$ & Depth $\downarrow$ & Normals $\downarrow$ & $\Delta_\text{MTL}$ (\%) $\uparrow$ & Flops (G) & MR$\downarrow$ \\
\midrule
STL & 43.20 & 0.599 & \textbf{19.42} & 0 & 1149 & 9.0 \\
MTL (Uni.) & 43.39 & 0.586 & 21.70 & -3.04 & 683 & 9.7 \\
DWA & 43.60 & 0.593 & 21.64 &  -3.16 & 683 & 9.7 \\
Uncertainty & 43.47 & 0.594 & 21.42 & -2.95 & 683 & 10.0 \\
Auto-$\lambda$ & 43.57 & 0.588 & 21.75 & -3.10 & 683 & 10.0 \\

RLW & 43.49 & 0.587 & 21.54 & -2.74 & 683 & 8.3 \\
\midrule
PCGrad & 43.74 & 0.588 & 21.55 & -2.66 & 683 & 7.3 \\
CAGrad & 43.57 & 0.583 & 21.55 & -2.49 & 683 & 7.0 \\
MGDA-UB & 42.56 & 0.586 & 21.76 & -3.83 & 683 & 11.3 \\
\midrule
MTAN & \textbf{44.92} & 0.585 & 21.14 & -0.84 & 683 & 4.0 \\
Cross-stitch & 44.19 & 0.577 & 19.62 & +1.66 & 1151 & 2.7 \\

\midrule
\methodname & 44.38 & \textbf{0.576} & 19.50 & \textbf{+2.04} & 916  & \textbf{1.7} \\
\methodname & 43.63 & 0.577 & 19.66 & +1.16 & 892 & 3.7 \\
\methodname & 43.05 & 0.589 & 19.95 & -0.50 & 794 & 9.7 \\
\bottomrule
\end{tabular*}
\end{sc}
\end{small}
\end{center}
\vskip -0.1in
\end{table}

\hyperref[table:nyud3]{Table~\ref{table:nyud3}} and \ref{table:nyud4} report the performance of DPT trained models with the ViT-base and ViT-small backbones. The MTL uniform and MTO baselines, display reduced computational cost, yet once again manifesting a performance drop in the normals prediction task. Similar to the trend between HRNet-18 and ResNet-50, the performance drop is more substantial for the smaller model, ViT-small, indicating that task interference is more prominent in small capacity settings. In comparison, \methodname{} consistently demonstrates a more favorable balance between the computational cost and the overall MTL accuracy across varied backbones.

\setlength{\tabcolsep}{+0.5 pt}

\begin{table}[!htb]
\caption{Results on NYUD-v2 using DPT with ViT-base.}
\label{table:nyud3}
\begin{center}
\begin{small}
\begin{sc}
\tiny
\begin{tabular*}{\columnwidth}{@{\extracolsep{\fill}}lcccccccc}
\toprule
Model & Semseg $\uparrow$ & Depth $\downarrow$ & Normals $\downarrow$ & $\Delta_\text{MTL}$ (\%) $\uparrow$ & Flops (G) & MR$\downarrow$ \\
\midrule
STL & 51.65 & 0.548 & \textbf{19.04} & 0 & 759& 5.0 \\
MTL (Uni.) & 51.38 & 0.539 & 20.73 & -2.57 & 294 & 7.3 \\
DWA & 51.66 & 0.536 & 20.98 & -2.66 & 294 & 6.0 \\
Uncertainty & 51.87 & 0.5352 & 20.72 & -2.02 & 294 & 4.0 \\
\midrule
\methodname & \textbf{51.98} & \textbf{0.528} & 19.10 & \textbf{+1.32} & 626 & \textbf{1.3} \\
\methodname & 51.46 & 0.536 & 19.34 & +0.08 & 483 & 5.0 \\
\methodname & 51.66 & 0.534 & 20.16 & -1.10 & 387 & 3.7 \\

\methodname & 51.71 & 0.535 & 20.38 & -1.51 & 324 & 3.7 \\
\bottomrule
\end{tabular*}
\end{sc}
\end{small}
\end{center}
\vskip -0.2in
\end{table}

\begin{table}[!htb]
\caption{Results on NYUD-v2 using DPT with ViT-small.}
\label{table:nyud4}
\begin{center}
\begin{small}
\begin{sc}
\tiny
\begin{tabular*}{\columnwidth}{@{\extracolsep{\fill}}lcccccccc}
\toprule
Model & Semseg $\uparrow$ & Depth $\downarrow$ & Normals $\downarrow$ & $\Delta_\text{MTL}$ (\%) $\uparrow$ & Flops (G) & MR$\downarrow$ \\
\midrule
STL & \textbf{46.58} & 0.583 & 21.22 & 0 & 248 & 4.0 \\
MTL (Uni.) & 45.32 & 0.576 & 22.86 & -3.04 & 118 & 7.3 \\
DWA & 45.74 & 0.5721 & 22.94 & -2.68 & 118 & 5.7 \\
Uncertainty & 45.67 & 0.5737 & 22.80 & -2.60 & 118 & 5.7 \\

\midrule

\methodname & 45.96 & \textbf{0.5648} & \textbf{20.77} & \textbf{+1.30} & 229 & \textbf{1.7} \\
\methodname & 45.34 & 0.5671 & 20.96 & +0.43 & 183 & 4.0 \\
\methodname & 45.57 & 0.5666 & 21.36 & +0.00 & 168 & 4.0 \\
\methodname & 45.99 & 0.5713 & 22.02 & -1.01 & 132 & 3.7 \\
\bottomrule
\end{tabular*}
\end{sc}
\end{small}
\end{center}
\vskip -0.2in
\end{table}

\subsubsection{PASCAL-Context}
\hyperref[table:pascal]{Table~\ref{table:pascal}} summarizes the results of our experiments on the PASCAL-context dataset encompassing five tasks.
Note that following previous work, we use the task losses'  weights $\omega_t$ from \cite{maninis2019attentive} for all MTL methods, but also report MTL uniform results as a reference.
\hyperref[fig:all_curves]{Figure~\ref{fig:all_curves}c}  illustrates the trade-off between $\Delta_\text{MTL}$ and the computational cost of all models.
The STL baseline outperforms most methods on the semantic segmentation and normals prediction tasks with a score of 14.70 and 66.1, while incurring a computational cost of 670 GFlops. Among the baseline MTL and MTO approaches, there is a notable degradation in surface normal prediction. Finally, as witnessed in prior works \cite{maninis2019attentive, vandenhende2020mti, Bruggemann_2021_ICCV}, we observe that most MTL and MTO baselines struggle to reach STL performance.
Among competing methods, 
MTAN and MGDA-UB yield the best MTL performance versus computational cost trade-off, however, both suffer from a notable decline in normals prediction performance.

At its highest compute budget (no sparsity loss and negligible computational savings), \methodname{} outperforms the STL baseline, notably in Saliency and Human parts prediction tasks, and achieves an overall $\Delta_\text{MTL}$ of +0.56. 
As we reduce the computational cost by increasing the sparsity loss weight $\lambda_s$, we observe a graceful decline in the multi-task performance that outperforms competing methods. Our \methodname{} models additionally obtain more favorable MR scores compared to the baselines. This emphasizes our model's ability to maintain a favorable balance between compute cost and multi-task performance across computational budgets.

\setlength{\tabcolsep}{+0.1 pt}
\vskip -0.1in
\begin{table}[!htb]
\caption{Performance comparison on PASCAL-Context.}
\label{table:pascal}

\begin{center}
\begin{small}
\tiny
\centering
\begin{tabular*}{\columnwidth}{@{\extracolsep{\fill}}lcccccccc}
\toprule
Model & Semseg $\uparrow$ & Normals $\downarrow$ & Saliency $\uparrow$ & Human $\uparrow$ & Edge $\downarrow$ & $\Delta_\text{MTL}$(\%) $\uparrow$ & Flops (G) & MR$\downarrow$ \\
\midrule
STL & 66.1 & 14.70 & 0.661 & 0.598 & 0.0175 & 0 & 670 & 6.0 \\
MTL (uniform) & 65.8 & 17.03 & 0.641 & 0.594 & 0.0176 & -4.14 & 284  & 12.0 \\
MTL (Scalar) & 64.3 & 15.93 & 0.656 & 0.586 & 0.0172 & -2.48 & 284  & 10.6 \\
DWA & 65.6 & 16.99 & 0.648 & 0.594 & 0.0180 & -3.91 & 284  & 12.0 \\
Uncertainty & 65.5 & 17.03 & 0.651 & 0.596 & 0.0174 & -3.68 & 284 & 10.2 \\
RLW & 65.2 & 17.22 & 0.660 & \textbf{0.634} & 0.0177 & -2.87 & 284  & 9.2 \\
\midrule
PCGrad & 62.6 & 15.35 & 0.645 & 0.596 & 0.0174 & -2.58 & 284  & 12.0 \\
CAGrad & 62.3 & 15.30 & 0.648 & 0.604 & 0.0174 & -2.03 & 284  & 10.2 \\
MGDA-UB & 63.0 & 15.34 & 0.646 & 0.604 & 0.0174 & -1.94 & 284 & 10.2 \\
\midrule
Cross-stitch & \textbf{66.3} & 15.13 & \textbf{0.663} & 0.602 & 0.0171 & +0.14 & 670   & 4.0 \\
MTAN & 65.1 & 15.76 & 0.659 & 0.590 & \textbf{0.0170} & -1.78 & 319   & 9.0 \\
\midrule
\methodname & 65.7 & 14.71 & \textbf{0.663} & 0.606 & 0.0172 & \textbf{+0.56} & 664  & \textbf{3.2} \\

\methodname & 65.1 & \textbf{14.64} & \textbf{0.663} & 0.604 & 0.0172 & +0.42 & 577   & 4.8 \\

\methodname & 65.2 & 14.75 & \textbf{0.663} & 0.600 & 0.0172 & +0.12 & 435  & 5.4 \\

\methodname & 64.9 & 14.72 & 0.658 & 0.596 & 0.0172 & -0.28 & 377  & 7.6 \\

\methodname & 65.1 & 15.02 & 0.655 & 0.592 & 0.0172 & -0.85 & 334  & 8.8 \\
\bottomrule

\end{tabular*}
\end{small}
\end{center}
\vskip -0.2in
\end{table}

\begin{table}[!htb]
\caption{Comparing the MTL performance using the $L_1$ Hinge loss and the standard $L_1$ loss on PASCAL-Context.}
\label{table:sparsity_ablation}
\vskip 0.1in
\begin{center}
\begin{small}
\tiny
\begin{tabular*}{\columnwidth}{@{\extracolsep{\fill}}lccccccccc}
\toprule
Model & $\mathcal{L_\text{sparsity}}$ & Semseg $\uparrow$ & Normals $\downarrow$ & Saliency $\uparrow$ & Human $\uparrow$ & Edge $\downarrow$ & $\Delta_\text{MTL}$(\%) $\uparrow$ & Flops (G) & MR$\downarrow$ \\
\midrule
\methodname & None   & 65.7 & 14.71 & 0.663 & 0.606 & 0.0172 & +0.56 & 664 & 1.8 \\
\midrule
\methodname &  $L_1$ & 63.9 & 14.74 & 0.664 & 0.600 & 0.0172 & -0.27 & 623 & 2.8 \\
\methodname &  $L_1$ & 61.1 & 15.07 & 0.663 & 0.582 & 0.0172 & -2.20 & 518 & 4.0 \\
\midrule
\methodname & Hinge & 65.1 & 14.64 & 0.648 & 0.604 & 0.0171 & +0.28 & 557 & 2.2 \\
\methodname & Hinge & 65.2 & 14.75 & 0.644 & 0.600 & 0.0172 & -0.13 & 433 & 3.4 \\
\bottomrule
\end{tabular*}
\end{small}
\end{center}
\vskip -0.1in
\end{table}

\subsection{Ablation Studies}
\label{ablations}

\subsubsection{Sparsity Loss}
To study the effect of the sparsity loss defined in \hyperref[eq:sparsity]{Equation \ref{eq:sparsity}}, we conduct the following two experiments:
First, we omit the sparsity regularization loss ($\lambda_s = 0$): As can be seen in the first row of \hyperref[table:sparsity_ablation]{Table~\ref{table:sparsity_ablation}}, \methodname{} outperforms the single task baseline,  but the computational savings are very limited.
In the second ablation experiment, we compare the use of the  $L_1$ hinge loss with a standard $L_1$ loss function as the sparsity regularizer: The results of \hyperref[table:sparsity_ablation]{Table~\ref{table:sparsity_ablation}} show that the hinge loss formulation consistently yields better trade-offs.

\subsubsection{Learned Sharing and Specialization Patterns}
\label{gatingpattern}

\begin{figure*}[!htbp]
\centering
\includegraphics[width=\textwidth]{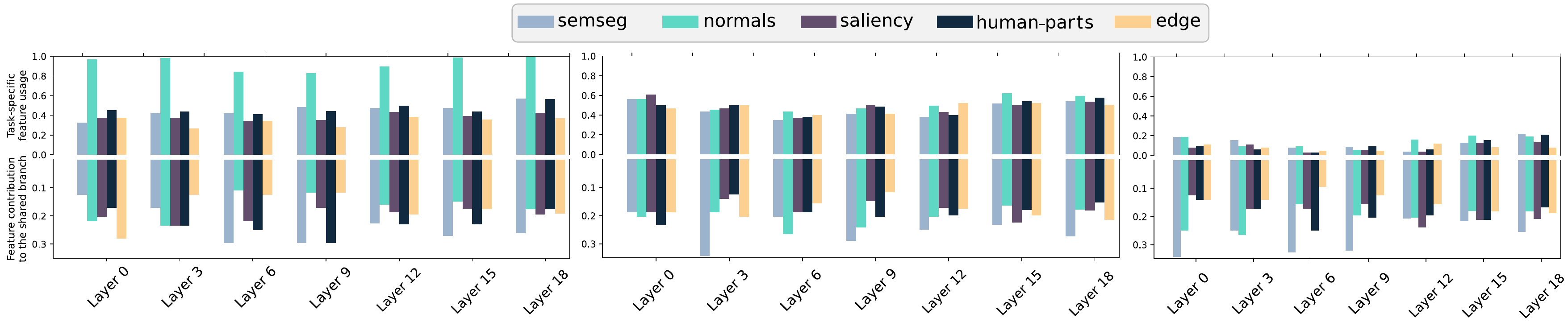}
\caption{The task-specific representation selection ratio (top) versus proportions of maximum contributions to the shared branch (bottom) for \methodname{} with hinge loss (left), $L_1$ loss with medium pruning (middle) and $L_1$ loss with high pruning (right).}
\label{fig:both_figures}
\vskip -0.1in
\end{figure*}

We then investigate the gating patterns that the model converges to.
Specifically, we want to observe how much each task contributes to and benefits from the shared representations. To that aim, we monitor \textbf{(i)} the percentage of task-specific representations selected by each task (captured by the gates $G_t(\alpha_t)$), as well as \textbf{(ii)} how much the features specific to each task contribute to the formation of the shared feature bank (captured by the learned combination weights $\beta$);
We visualize these values for the five tasks of the Pascal-Context dataset in \hyperref[fig:both_figures]{Figure~\ref{fig:both_figures}} for different sparsity regularizers: with hinge loss (\textit{left}), with $L_1$ loss at medium (middle) and high pruning levels (\textit{right}). 
We also present these results for all layers in \hyperref[fig:gating_dist_full]{Figure~\ref{fig:gating_dist_full}}, \hyperref[app:results]{Appendix \ref{app:results}}.

In all settings, the semantic segmentation task makes the largest contribution to the shared branch, followed by the normals prediction tasks.
It is worth noting that the amount of feature contribution to the shared branch can also be largely influenced by other tasks' loss functions. In this situation, we observe that if the normal task lacks enough task-specific features (as seen in the middle and right models), its performance deteriorates significantly. In contrast, when it acquires sufficient task-specific features, it maintains a high accuracy (\textit{left}). Intriguingly, the features of the normal task become less interesting to other tasks in this scenario possibly due to increased specialization.

\subsubsection{Impact of Model Capacity}
\label{cap_ablation}
In this section, we conduct an ablation study to analyze the relationship between model capacity and multi-task performance. We progressively reduce the width of ResNet-50 and ResNet-20 to half and a quarter of the original sizes for NYUD-v2 and CelebA datasets, respectively. Shrinking the model size, as observed in \hyperref[table:capacity1]{Table \ref{table:capacity1}}, incurs progressively more harmful effect on multi-task performance compared to the single task baseline. In comparison, our proposed \methodname{} approach consistently finds a favorable trade-off between capacity and performance and improves over single task performances, across all capacity ranges.

\section{Discussion and Conclusion}
\label{sec:discussion}
In this paper, we propose \methodname, a novel framework to address the fundamental challenges of task interference and computational constraints in MTL. \methodname{} leverages a learnable gating mechanism that enables individual tasks to select and combine channels from both a specialized and shared feature set. This framework promotes an asymmetric flow of information, where each task is free to decide how much it contributes to -- or takes from -- the shared branch. 
By regularizing the learnable gates, we can strike a balance between task-specific resource allocation and overarching computational costs. \methodname{} demonstrates state-of-the-art performance across various architectures and on notable benchmarks such as CelebA, NYUD-v2, and Pascal-Context. The gating mechanism in \methodname{} operates over the channel dimension, or the embedding dimension in the case of ViTs, which in its current form does not support structured pruning for attention matrix computations. Future explorations might integrate approaches like patch-token gating to further optimize computational efficiency.
\setlength{\tabcolsep}{+2 pt}
\begin{table}[!h]
\caption{Performance across various model capacities using the ResNet-20 and ResNet-50 backbones on the CelebA  \textit{(top)} and NYUD-v2 \textit{(bottom)} tasks.}
\vskip +0.05in
\begin{minipage}{\linewidth}
\label{table:capacity1}
\begin{center}
\begin{small}
\begin{sc}
\tiny
\begin{NiceTabular}{c|lcccccc}

\toprule
& Model & Gender $\uparrow$ & Age $\uparrow$ & Clothes $\uparrow$ & Overall $\uparrow$ & Flops (M) & MR$\downarrow$ \\
\midrule
\Block{3-1}<\rotate>{Original} & STL & 97.50 & 86.02 & \textbf{93.00} & 92.17 & 174& 2.0 \\
& MTL & 97.28 & 86.70 & 92.35 & 92.11 & 58 & 2.7 \\
& \methodname & \textbf{97.60} & \textbf{87.44} & 92.40 & \textbf{92.48} & 59 & \textbf{1.3} \\
\midrule
\Block{3-1}<\rotate>{Half} & STL & 96.99 & 85.60 & \textbf{92.72} & 91.77 & 44.4 & 2.3 \\
& MTL & 97.02 & 86.41 & 92.11 & 91.85 & 14.8& 2.0 \\
& \methodname & \textbf{97.33} & \textbf{86.75} & 92.05 & \textbf{92.05} & 15.5& \textbf{1.7} \\
\midrule
\Block{3-1}<\rotate>{Quarter} & STL & 96.64 & 85.22 & \textbf{92.19} & 91.35 & 11.6 & 2.0 \\
& MTL & 96.46 & 85.46 & 91.59 & 91.17 & 3.9 & 2.3 \\
& \methodname & \textbf{96.81} & \textbf{86.05} & 91.48 & \textbf{91.45} & 4.7& \textbf{1.7} \\

\bottomrule

\end{NiceTabular}
\end{sc}
\end{small}
\end{center}
\end{minipage}

\vskip 0.1in
\begin{minipage}{\linewidth}
\label{table:capacity2}
\begin{center}
\begin{small}
\begin{sc}
\tiny
\setlength{\tabcolsep}{+1.1 pt}
\begin{NiceTabular}{c|lcccccc}

\toprule
& Model & Semseg $\uparrow$ & Depth $\downarrow$ & Normals $\downarrow$ & $\Delta_\text{MTL}$ (\%) $\uparrow$ & Flops (G) & MR$\downarrow$ \\
\midrule
\Block{3-1}<\rotate>{Original} & STL & 43.20 & 0.599 & \textbf{19.42} & 0 & 1149 & 2.3 \\
& MTL & 43.39 & 0.586 & 21.70 & -3.02 & 683 & 2.3 \\
& \methodname & \textbf{43.63} & \textbf{0.577} & 19.66 & \textbf{+1.16}  & 892 & \textbf{1.3} \\
\midrule
\Block{3-1}<\rotate>{Half} & STL & 39.72 & 0.613 & \textbf{20.06} & 0 & 415 & 2.3 \\
& MTL & 40.20 & 0.610 & 22.78 & -3.98 & 296 & 2.0 \\
& \methodname & \textbf{39.78} & \textbf{0.591} & 20.41 & \textbf{+0.63}  & 348 & \textbf{1.7} \\
\midrule
\Block{3-1}<\rotate>{Quarter} & STL & 35.44 & 0.654 & \textbf{21.21} & 0 & 177 & 2.3 \\
& MTL & 35.68 & 0.632 & 24.57 & -4.06 & 147 & 2.3 \\
& \methodname & \textbf{35.71} & \textbf{0.624} & 21.75 & \textbf{+0.94}  & 164 & \textbf{1.3} \\

\bottomrule
\end{NiceTabular}
\end{sc}
\end{small}
\end{center}
\end{minipage}
\end{table}

\textbf{Limitations.} In this work, we primarily focus on enhancing the trade-off between accuracy and efficiency during inference. However, \methodname{} comes with a moderate increase in training time.
While we observe that \methodname{} demonstrates SoTA results on PASCAL-Context with five tasks, further evaluation on MTL setups with larger number of tasks remains to be explored.
Furthermore, although both $\lambda_s$ and $\tau_t$ can control the trade-off between performance and computational cost, effectively approximating the desired FLOPs, we still cannot guarantee a specific target FLOP. 

\section{Impact Statement}
This paper presents work whose goal is to advance the field of Machine Learning. There are many potential societal consequences of our work, none which we feel must be specifically highlighted here.
\bibliography{icml2024}
\bibliographystyle{icml2024}
\newpage
\appendix
\onecolumn
\clearpage
\setcounter{section}{0}
\section*{Appendix}
\section{Implementation Details}
\label{Implementation}

\subsection{CelebA}

On the CelebA dataset, we use  ResNet-20 as our backbone with three task-specific linear classifier heads, one for each attribute. We resize the input images to 32x32 and remove the initial pooling in the stem of ResNet to accommodate the small image resolution.
For training, we use the Adam optimizer with a learning rate of 1e-3, weight decay of 1e-4, and a batch size of 128. For learning rate decay, we use a step learning rate scheduler with step size 20 and a multiplicative factor of \(1/3\). We use SGD with a learning rate of 0.1 for the gates' parameters.

\subsection{NYUD and PASCAL-Context}
For both NYUD-v2 and PASCAL-Context with ResNet-18 and ResNet-50 backbones, we use the Atrous Spatial Pyramid Pooling (ASPP) module introduced by \cite{chen2018encoder} as task-specific heads.
For the HRNet-18 backbone, we follow the methodology of the original paper \cite{wang2020deep}: HRNet  combines the output representations at four different resolutions and fuses them using 1x1 convolutions to output dense prediction.

We train all convolution-based encoders on the NYUD-v2 dataset for 100 epochs with a batch size of 4 and on the PASCAL-Context dataset for 60 epochs with a batch size of 8. We use the Adam optimizer, with a learning rate of 1e-4 and weight decay of 1e-4. We use the same data augmentation strategies for both NYUD-v2 and PASCAL-Context datasets as described in \cite{vandenhende2020mti}. We use SGD with a learning rate of 0.1 to learn the gates' parameters.

In terms of task objectives, we use the cross-entropy loss for semantic segmentation and human parts, $L_1$ loss for depth and normals, and binary-cross entropy loss for edge and saliency detection tasks, similar to \cite{vandenhende2020mti}. For learning rate decay,  we adopt  a polynomial learning rate decay scheme with a power of 0.9.

\textbf{The Choice of $\omega_t$.} The hyper-parameter $\omega_t$ denotes the scalarization weights. We use the weights suggested in prior work but also report numbers of uniform scalarization. For NYUD-v2, we use uniform scalarization as suggested in \cite{maninis2019attentive,vandenhende2021multi}, and for PASCAL-Context, we similarly use the weights suggested in \cite{maninis2019attentive} and \cite{vandenhende2021multi}.

\subsection{DPT Training}

For DPT training, we follow the same training procedure as described by the authors, which employs the Adam optimizer, with a learning rate of 1e-5 for the encoder and 1e-4 for the task heads, and a batch size of 8.

The ViT backbones were pre-trained on ImageNet-21k at resolution 224$\times$224, and fine-tuned on ImageNet 2012 at resolution 384$\times$384.  The feature dimension for DPT's task heads was reduced from 256 to 64. We conducted a sweep over a set of weight decay values and chose 1e-6 as the optimal value for our DPT experiments.

\section{Forward-pass Pseudo-code}
\label{alg:cap}
Algorithm~\ref{alg:forwardpass} illustrates the steps in the forward pass of the algorithm.

\begin{algorithm}[!htbp]
\caption{Pseudo-code for unified representation encoder}
\label{alg:forwardpass}
\renewcommand{\algorithmiccomment}[1]{\hfill\eqparbox{COMMENT}{#1}}
\begin{algorithmic}
\STATE {\bfseries Given:}
\vspace{-\topsep}
\begin{itemize}
\setlength{\parskip}{0pt}
\setlength{\itemsep}{0pt}
\item $x \in \mathbb{R}^{3 \times W \times H}$ \COMMENT{Input image}
\item $T, L \in \mathbb{R}$ \COMMENT{Number of tasks and encoder layers}
\item $\shrparams$, $\taskparams$ \COMMENT{Shared and $t$-th task-specific layer parameters}
\item $\beta$, $\alpha_t$     \COMMENT{Shared and $t$-th task-specific gating parameters}
\end{itemize}
\vspace{-\topsep}
\STATE {\bfseries Return:} [$\ensuremath{\varphi_1}^L,..., \ensuremath{\varphi_T}^L$]  \COMMENT{Task-specific encoder representations}
\STATE $\shrfeat^0, \ensuremath{\varphi_1}^0,..., \ensuremath{\varphi_T}^0  \gets x$ \COMMENT{Set initial shared and task-specific features}
\FOR{$\ell=1$ to $L$}
\FOR{$t=1$ to $T$}
\STATE $\varphi^{\prime \ell}_t \gets G_t^\ell (\alpha_t^\ell) \odot \taskfeat^\ell + (1 - G_t^\ell (\alpha_t^\ell)) \odot \shrfeat^\ell$ (\ref{eq:taskmix})  \COMMENT{Choose shared and task-specific features}
\STATE $\taskfeat^{\ell+1} \gets F(\taskfeat^{\prime \ell}; \taskparams^\ell)$  \COMMENT{Compute task-specific features}
\ENDFOR
\STATE $\shrfeat^{\prime \ell} = \sum_{t=1}^{T} \underset{t=1 \dots T}{\text{softmax}}(\combwgt^\ell) \odot \taskfeat^{\prime \ell}$ (\ref{shrmix}) \COMMENT{Combine task-specific features to form shared ones}
\STATE $\shrfeat^{\ell+1} \gets F(\shrfeat^{\prime \ell}; \shrparams^\ell)$ \COMMENT{Compute shared features}
\ENDFOR
\end{algorithmic}
\end{algorithm}

\section{Generalization to Vision Transformers}
\label{methodvit}
As transformers are becoming widely used in the vision literature, and to show the generality of our proposed MTL framework, we also apply \methodname{} to vision transformers:
We again denote $\taskfeat^\ell \in \mathbb{R}^{ N^\ell \times C^\ell}$ and $\shrfeat^\ell \in \mathbb{R}^{ N^\ell \times C^\ell}$ as the $t$-th task-specific and shared representations in layer $\ell$, where $N^\ell$ and $C^\ell$ are the number of tokens and embedding dimensions, respectively.
We first apply our feature selection to the key, query and value linear projections in each self-attention block:
\begin{align}
q_t^{l+1} &= G_t^\ell (\gateparams_{q,t}^\ell) \odot f_{q,t}^\ell(\taskfeat^\ell; \taskparams^\ell) + (1 - G_t^\ell (\gateparams_{q,t}^\ell)) \odot f_q^\ell(\shrfeat^\ell; \shrparams^\ell),\\
k_t^{l+1} &= G_t^\ell (\gateparams_{k,t}^\ell) \odot f_{k,t}^\ell(\taskfeat^\ell; \taskparams^\ell) + (1 - G_t^\ell (\gateparams_{k,t}^\ell)) \odot f_k^\ell(\shrfeat^\ell; \shrparams^\ell),\\
v_t^{l+1} &= G_t^\ell (\gateparams_{v,t}^\ell) \odot f_{v,t}^\ell(\taskfeat^\ell; \taskparams^\ell) + (1 - G_t^\ell (\gateparams_{v,t}^\ell)) \odot f_v^\ell(\shrfeat^\ell; \shrparams^\ell),
\end{align}
where $\gateparams_{q,t}^\ell$, $\gateparams_{k,t}^\ell$, $\gateparams_{v,t}^\ell$ are the learnable gating parameters mixing the task-specific and shared projections for queries, keys and values, respectively. $f_{q, t}^l$, $f_{k, t}^l$, $f_{v, t}^l$ are the linear projections for query, key and value for the task $t$, while $f_{q}^l$, $f_{k}^l$, $f_{v}^l$ are the corresponding shared projections.
Once the task-specific representations are formed, the shared embeddings for the next block are computed by a learned mixing of the task-specific feature followed by a linear projection, as described in \hyperref[shrmix]{(\ref{shrmix})}.
Similarly, we apply this gating mechanism to the final linear projection of the multi-head self-attention, as well as the linear layers in the feed-forward networks in-between each self-attention block.

\section{Additional Experiments}
\label{app:results}
\subsection{Full Results on CelebA and NYUD-v2}
In \hyperref[table:full_celebA]{Table~\ref{table:full_celebA}}, we report results on the CelebA dataset for different model capacities: Here,  \methodname{} is  compared to the STL and standard MTL methods with different model  width: at original, half and  quarter of the original model width. In  \hyperref[table:nyud1a]{Tables~\ref{table:nyud1a}} and \hyperref[table:nyud2a]{\ref{table:nyud2a}}, we report the complete results on NYUD-v2 dataset using HRNet-18 and ResNet-50 backbones, including the parameter count and standard deviation of the $\Delta_\text{MTL}$ scores.

\subsection{Sharing/Specialization Patterns}

\begin{table}[h]
\scriptsize
\centering
\caption{Performance comparison of various MTL models on the CelebA dataset with different model capacities. Different \methodname{} models are obtained by varying $\lambda_s$}.
\label{table:full_celebA}

\begin{NiceTabular}{c|lcccccc}

\toprule
& Model & Gender $\uparrow$ & Age $\uparrow$ & Clothes $\uparrow$ & Overall $\uparrow$ & Flops (M) & MR$\downarrow$ \\
\midrule
\Block{3-1}<\rotate>{Original} & STL & 97.50 & 86.02 & \textbf{93.00} & 92.17 & 174& 3.3 \\
& MTL & 97.28 & 86.70 & 92.35 & 92.11 & 58 & 4.7 \\

& \methodname & 97.60 & 87.44 & 92.40 & 92.48 & 59 & 2.7 \\
& \methodname & 97.77 & 87.39 & 92.56 & 92.57 & 11 & 2.3 \\
& \methodname & \textbf{97.95} & \textbf{87.24} & 92.85 & \textbf{92.68} & 162 & \textbf{2.0} \\

\midrule
\Block{3-1}<\rotate>{Half} & STL & 96.99 & 85.60 & \textbf{92.72} & 91.77 & 44.4 & 3.7 \\
& MTL & 97.02 & 86.41 & 92.11 & 91.85 & 14.8 & 4.0 \\
& \methodname & 97.33 & 86.75 & 92.05 & 92.05 & 15.5 & 3.7 \\
& \methodname & 97.33 & \textbf{87.05} & 92.12 & 92.17 & 17.4 & 2.3 \\
& \methodname & \textbf{97.46} & 86.97 & 92.47 & \textbf{92.23} & 24.5& \textbf{1.7} \\

\midrule
\Block{3-1}<\rotate>{Quarter} & STL & 96.64 & 85.22 & \textbf{92.19} & 91.35 & 11.6& 3.3 \\
& MTL & 96.46 & 85.46 & 91.59 & 91.17 & 3.9 & 4.3 \\
& \methodname & 96.81 & 86.05 & 91.48 & 91.45 & 4.7& 3.7 \\
& \methodname & \textbf{96.92} & 86.10 & 91.64 & 91.56 & 5.5& \textbf{2.0} \\
& \methodname & 96.81 & \textbf{86.61} & 91.74 & \textbf{91.72} & 6.4& \textbf{2.0} \\
\bottomrule

\end{NiceTabular}
\end{table}

\setlength{\tabcolsep}{+1 pt}

\begin{table}[!htb]
\caption{Performance comparison on NYUD-v2 using HRNet-18 backbone. Different \methodname{} models are obtained by varying $\lambda_s$.}
\label{table:nyud1a}
\vskip 0.1in
\begin{center}
\begin{small}
\begin{sc}
\tiny
\begin{tabular*}{0.5\columnwidth}{lcccccccc}
\toprule
Model & Semseg $\uparrow$ & Depth $\downarrow$ & Normals $\downarrow$ & $\Delta_\text{MTL}$ (\%) $\uparrow$ & Flops (G) & Param (M) & MR$\downarrow$ \\
\midrule
STL & 41.70 & 0.582 & \textbf{18.89} & 0 $\pm$ 0.12 & 65.1 & 28.9 & 8.0 \\
MTL (Uni.) & 41.83 & 0.582 & 22.84 & -6.86 $\pm$ 0.76 & 24.5 & 9.8 & 11.0 \\
DWA & 41.86 & 0.580 & 22.61 &  -6.29 $\pm$ 0.95 & 24.5 & 9.8 & 8.7 \\
Uncertainty & 41.49 & 0.575 & 22.27 & -5.73 $\pm$ 0.35 & 24.5 & 9.8 & 8.3 \\
Auto-$\lambda$ & 42.71 & 0.577 & 22.87 &  -5.92 $\pm$ 0.47 & 24.5 & 9.8 & 8.0 \\
RLW & 42.10 & 0.593 & 23.29 &  -8.09 $\pm$ 1.11 & 24.5 & 9.8 & 11.7 \\
\midrule
PCGrad & 41.75 & 0.581 & 22.73 &  -6.70 $\pm$ 0.99 & 24.5 & 9.8 & 10.3 \\
CAGrad & 42.31 & 0.580 & 22.79 &  -6.28 $\pm$ 0.90 & 24.5 & 9.8 & 8.7 \\
MGDA-UB & 41.23 & 0.625 & 21.07 &  -6.68 $\pm$ 0.67 & 24.5 & 9.8 & 11.3 \\

\midrule
\methodname & \textbf{43.58} & \textbf{0.559} & 19.32 & \textbf{+2.06} $\pm$ 0.13 & 43.2 & 18.8 & \textbf{1.3} \\
\methodname & 42.95 & 0.562 & 19.73 & +0.68 $\pm$ 0.09 & 38.3 & 16.5 & 2.3 \\
\methodname & 42.36 & 0.564 & 20.04 & -0.55 $\pm$ 0.17& 36.0 & 15.4 & 4.0 \\
\methodname & 42.73 & 0.575 & 21.01 & -2.55 $\pm$ 0.11 & 33.1 & 13.7 & 4.0 \\
\methodname & 42.35 & 0.575 & 21.70 & -4.07 $\pm$ 0.38 & 29.2  & 11.9 & 5.7 \\
\bottomrule
\end{tabular*}
\end{sc}
\end{small}
\end{center}
\vskip -0.1in
\end{table}

\begin{table}[!htb]
\caption{Performance comparison on NYUD-v2 using ResNet-50 backbone. Different \methodname{} models are obtained by varying $\lambda_s$.}
\label{table:nyud2a}
\vskip 0.1in
\begin{center}
\begin{small}
\begin{sc}
\tiny
\begin{tabular*}{0.5\columnwidth}{lcccccccc}
\toprule
Model & Semseg $\uparrow$ & Depth $\downarrow$ & Normals $\downarrow$ & $\Delta_\text{MTL}$ (\%) $\uparrow$ & Flops (G) & Param (M) & MR$\downarrow$ \\
\midrule
STL & 43.20 & 0.599 & \textbf{19.42} & 0 $\pm$ 0.11 & 1149 & 118.9 & 9.0 \\
MTL (Uni.) & 43.39 & 0.586 & 21.70 & -3.04 $\pm$ 0.79 & 683 & 71.9 & 9.7 \\
DWA & 43.60 & 0.593 & 21.64 &  -3.16 $\pm$ 0.39 & 683 & 71.9 & 9.7 \\
Uncertainty & 43.47 & 0.594 & 21.42 & -2.95 $\pm$ 0.40 & 683 & 71.9 & 10.0 \\
Auto-$\lambda$ & 43.57 & 0.588 & 21.75 & -3.10 $\pm$ 0.39 & 683 & 71.9 & 10.0 \\

RLW & 43.49 & 0.587 & 21.54 & -2.74 $\pm$ 0.09 & 683 & 71.9 & 8.3 \\
\midrule
PCGrad & 43.74 & 0.588 & 21.55 & -2.66 $\pm$ 0.15 & 683 & 71.9 & 7.3 \\
CAGrad & 43.57 & 0.583 & 21.55 & -2.49 $\pm$ 0.11 & 683 & 71.9 & 7.0 \\
MGDA-UB & 42.56 & 0.586 & 21.76 & -3.83 $\pm$ 0.17 & 683 & 71.9 & 11.3 \\
\midrule
MTAN & \textbf{44.92} & 0.585 & 21.14 & -0.84 $\pm$ 0.32 & 683 & 92.4 & 4.0 \\
Cross-stitch & 44.19 & 0.577 & 19.62 & +1.66 $\pm$ 0.09 & 1151 & 119.0 & 2.7 \\

\midrule
\methodname & 44.38 & \textbf{0.576} & 19.50 & \textbf{+2.04} $\pm$ 0.07 & 916  & 95.4 & \textbf{1.7} \\
\methodname & 43.63 & 0.577 & 19.66 & +1.16 $\pm$ 0.10 & 892 & 92.4 & 3.7 \\
\methodname & 43.05 & 0.589 & 19.95 & -0.50 $\pm$ 0.05 & 794 & 83.3 & 9.7 \\
\bottomrule
\end{tabular*}
\end{sc}
\end{small}
\end{center}
\vskip -0.1in
\end{table}

\begin{figure}[!htbp]
\centering

\begin{subfigure}[b]{\linewidth}
\includegraphics[width=\linewidth]{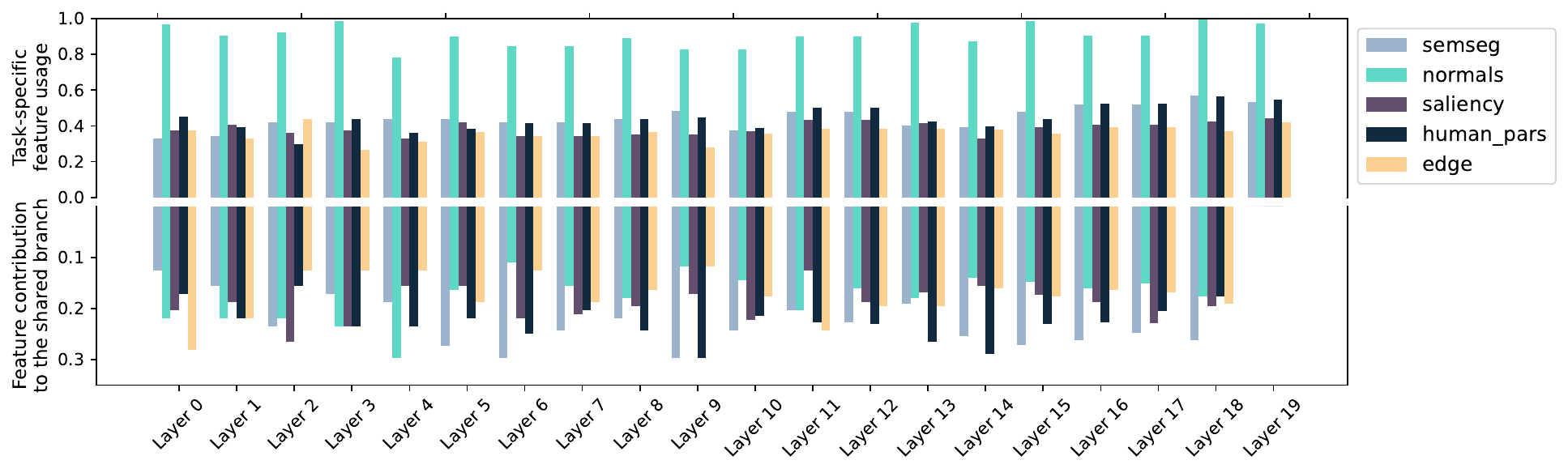}
\caption{With Hinge sparsity loss}
\label{fig:sub1}
\end{subfigure}

\begin{subfigure}[b]{\linewidth}
\includegraphics[width=\linewidth]{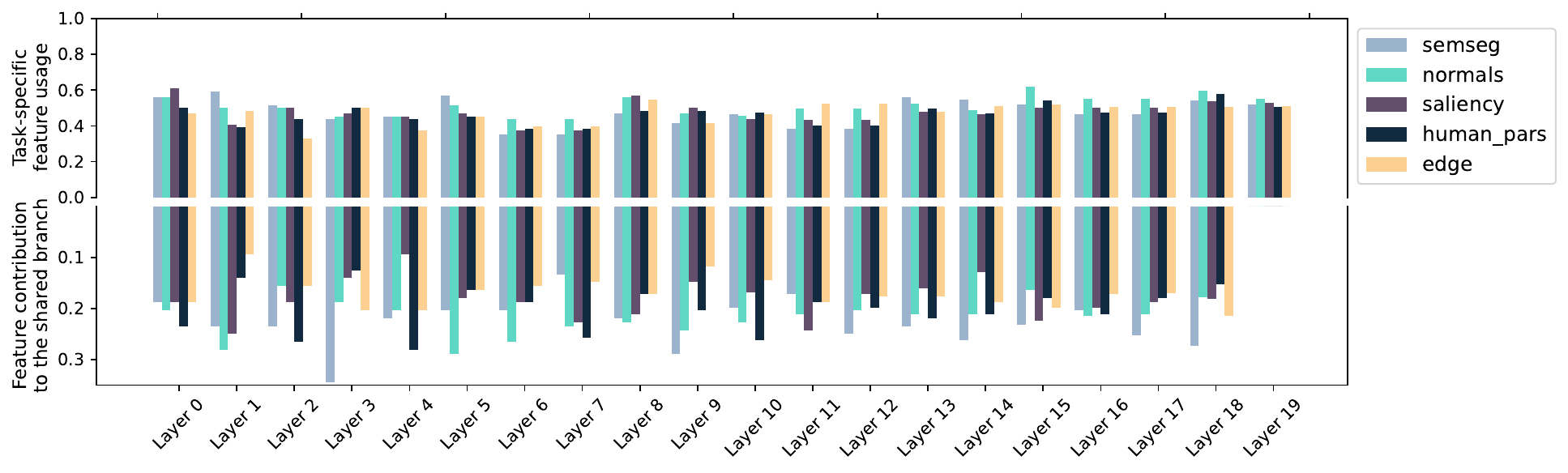}
\caption{With uniform $L_1$ loss (medium pruning)}
\label{fig:sub2}
\end{subfigure}

\begin{subfigure}[b]{\linewidth}
\includegraphics[width=\linewidth]{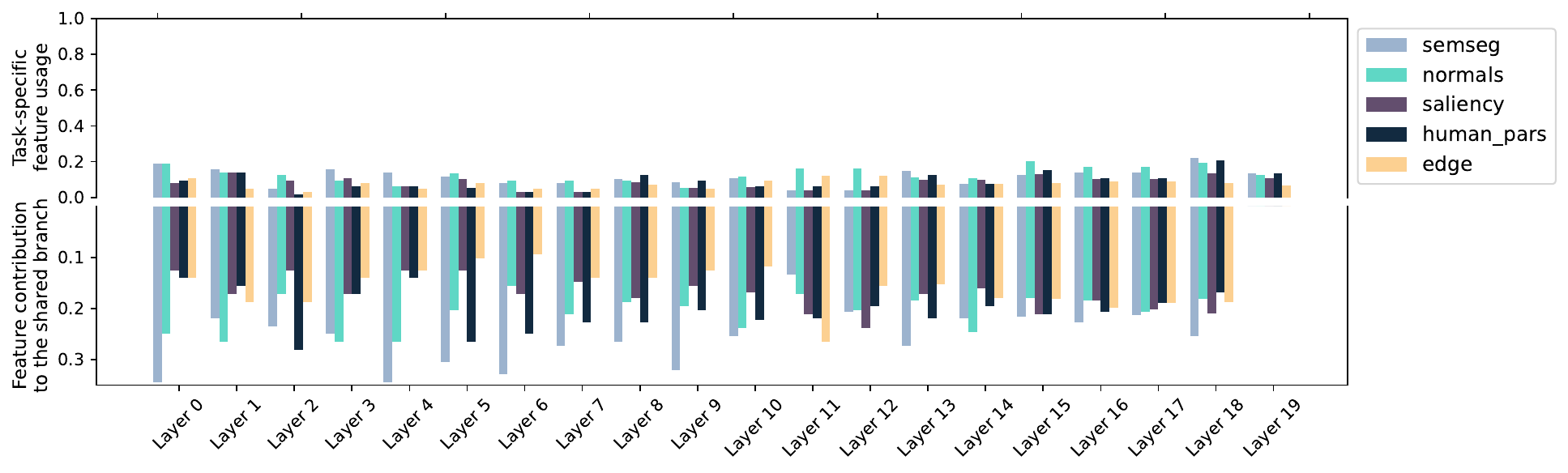}
\caption{With uniform $L_1$ loss (high pruning)}
\label{fig:sub3}
\end{subfigure}

\caption{Sharing and specialization patterns on pascal context dataset with ResNet-18 backbone.}
\label{fig:gating_dist_full}
\end{figure}

\hyperref[fig:gating_dist_full]{Figure~\ref{fig:gating_dist_full}} illustrates the distribution of the gating patterns across all layers of the ResNet-18 backbone for the PASCAL-Context dataset for 3 models using (a) a Hinge loss, (b) a medium-level pruning using uniform $L_1$ loss and (c) a high-level pruning with uniform $L_1$ loss.

\section{Ablation: Sparsity Targets}
\label{app:sparsity}
By tuning the sparsity targets $\tau$ in \hyperref[eq:sparsity]{Equation \ref{eq:sparsity}}, we can achieve specific compute budgets of the final network at inference.
However, there are multiple choices of $\{\tau_t\}_{t=1}^T$ that can achieve the same budget. In this section, we further investigate the impact of which task we allocate more or less of the compute budget on the final accuracy/efficiency trade-off.

We perform an experiment sweep for different combination of sparsity targets, where each $\tau_t$ is chosen from $\{0, 0.25, 0.75, 1.0\}$. Each experiment is run for two different random seeds and two different sparsity loss weights $\lambda_s$.
Due to the large number of experiments, we perform the ablation experiments for shorter training runs (75\% of the training epochs for each setup)

Our take-away conclusions are that \textbf{(i)} we clearly observe that some tasks require more task-specific parameters (hence a higher sparsity target) and \textbf{(ii)} this dichotomy often correlates with the per-task performance gap observed between the STL and MTL baselines, which can thus be used as a guide to set the hyperparameter values for $\tau$.

\begin{figure}[!htbp]
\centering
\begin{minipage}[c]{0.45\textwidth}
\centering
\includegraphics[width=\linewidth]{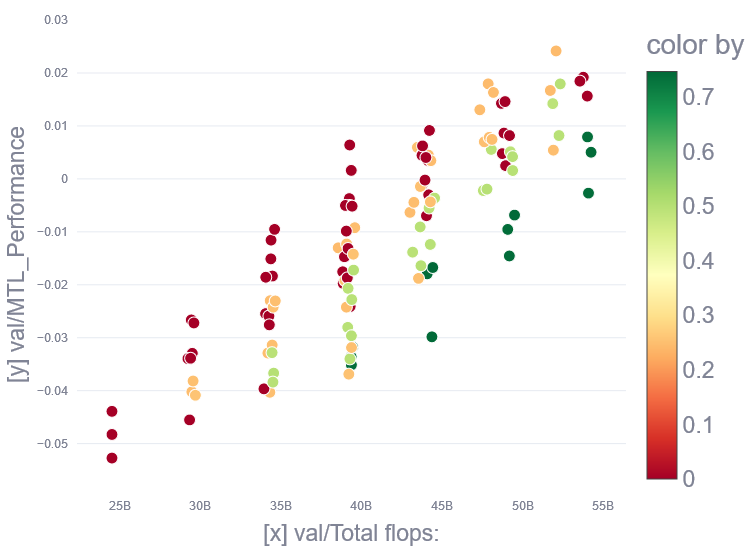}

(a) Color by $\tau_{\text{segmentation}}$
\end{minipage}~
\begin{minipage}[c]{0.45\textwidth}
\centering
\includegraphics[width=\linewidth]{ablation_tau/nyud_depth.png}

(b) Color by $\tau_{\text{depth}}$
\end{minipage}

\begin{minipage}[c]{0.45\textwidth}
\centering
\includegraphics[width=\linewidth]{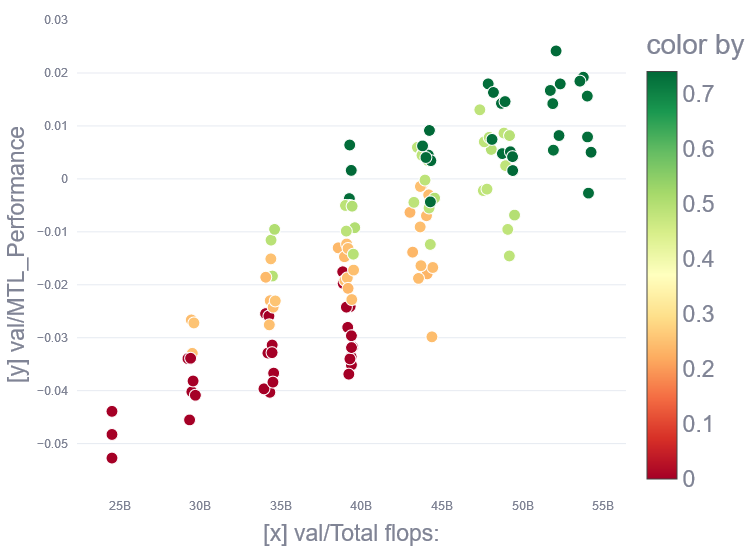}

(c) Color by $\tau_{\text{normals}}$
\end{minipage}
\caption{Sweeping over different $\{\tau_t\}$ on the NYUD-v2 experiments with HRNet-18 backbone. We plot the MTL performance $\Delta_{MTL}$ against the total number of FLOPs, then color each scatter point by the value of $\tau_t$ when the task $t$ is (a) segmentation, (b) depth and (c) normals.}
\label{fig:sparsity-nyu}
\end{figure}

In the results of NYUD-v2 in \hyperref[fig:sparsity-nyu]{Figure \ref{fig:sparsity-nyu}}, we observe a clear hierarchy in terms of task importance: When looking at the points on the Pareto curve, they prefer high values of $\tau_{\text{normals}}$, followed by $\tau_{\text{segmentation}}$: In other words, these two tasks, and in particular normals prediction, requires more task-specific parameters than the depth prediction task to obtain the best MTL performance versus compute cost trade-offs.

\begin{figure}[!htbp]
\centering
\begin{minipage}[c]{0.45\textwidth}
\centering
\includegraphics[width=\linewidth]{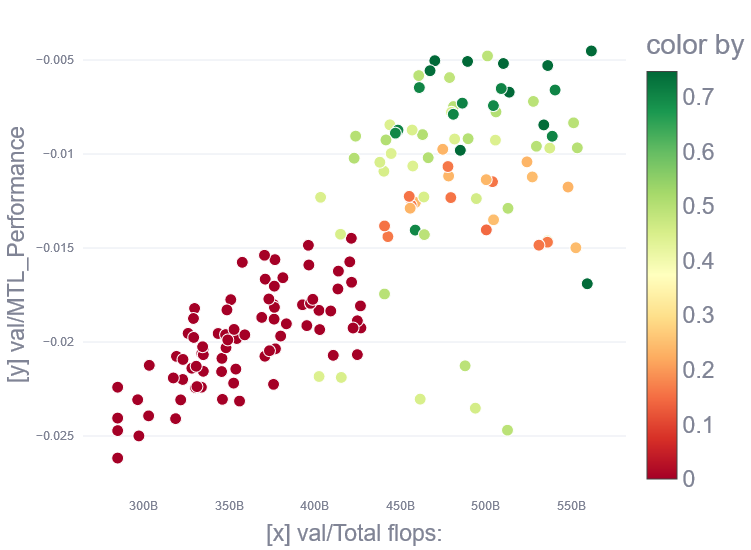}

(a) Color by $\tau_{\text{edges}}$
\end{minipage}~
\begin{minipage}[c]{0.45\textwidth}
\centering
\includegraphics[width=\linewidth]{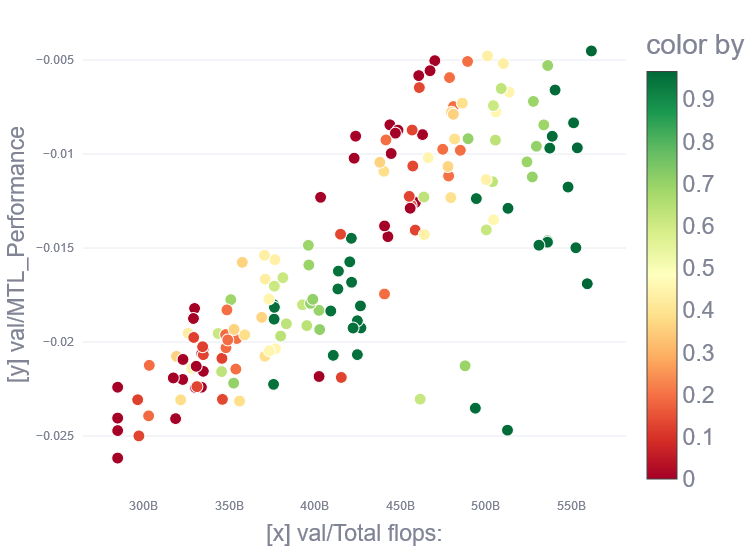}

(b) Color by $\tau_{\text{normals}}$
\end{minipage}

\begin{minipage}[c]{0.45\textwidth}
\centering
\includegraphics[width=\linewidth]{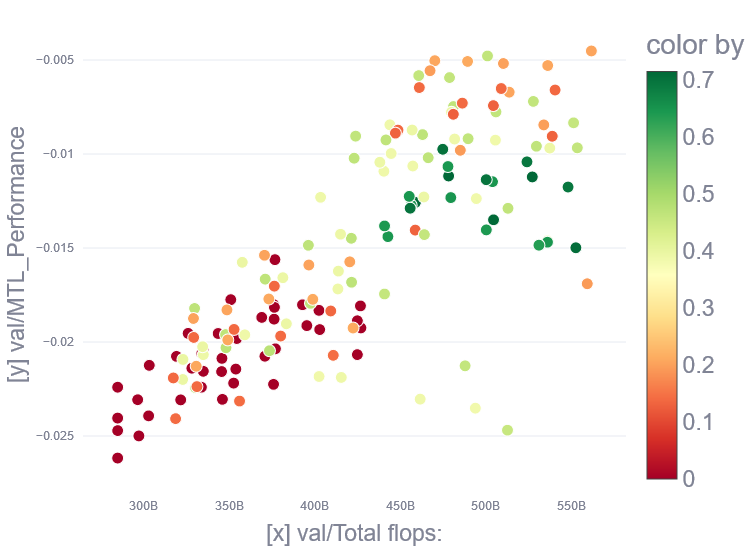}

(c) Color by $\tau_{\text{human parts}}$
\end{minipage}~
\begin{minipage}[c]{0.45\textwidth}
\centering
\includegraphics[width=\linewidth]{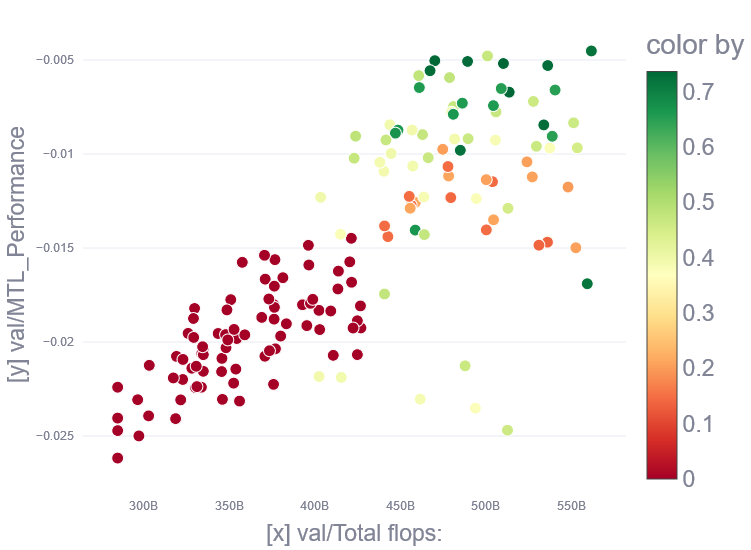}

(d) Color by $\tau_{\text{saliency}}$
\end{minipage}

\begin{minipage}[c]{0.45\textwidth}
\centering
\includegraphics[width=\linewidth]{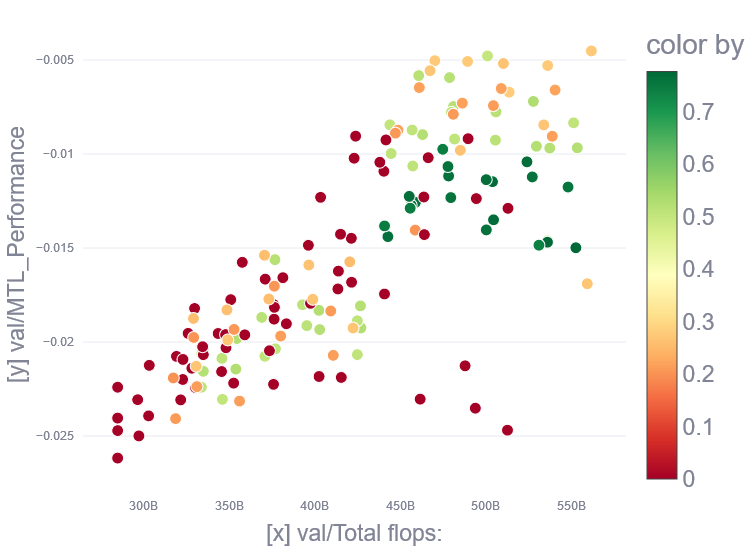}

(e) Color by $\tau_{\text{segmentation}}$
\end{minipage}
\caption{Sweeping over different $\{\tau_t\}$ on the PASCAL-Context. We plot the MTL performance $\Delta_{MTL}$ against the total number of FLOPs, then color each scatter point by the value of $\tau_t$ when the task $t$ is (a) edges, (b) normals (c) human parts, (d) saliency and (e) segmentation.}
\label{fig:sparsity-pascal}
\end{figure}

Then, we conduct a similar analysis for the five tasks of PASCAL-Context in \hyperref[fig:sparsity-pascal]{Figure~\ref{fig:sparsity-pascal}}.
Here we see a clear split in tasks: The graph for the edges prediction and saliency task are very similar to one another and tend to prefer high $\tau$ values, i.e. more task-specific parameters, at higher compute budget. But when focusing on a lower compute budget, it is more beneficial to the overall objective for these tasks to use the shared branch.
Similarly, the tasks of segmentation and human parts exhibit similar behavior under variations of $\tau$ and are more robust to using shared representations (lower values of $\tau$).
Finally, the task of normals prediction (b) differ from the other four, and in particular exhibit a variance of behavior across different compute budget. In particular, when targetting the intermediate range  (350B-450B FLOPs), setting higher $\tau_{\text{normals}}$ helps the overall objective.

\section{Training Time Comparisons}
\label{app:timing}
While our method is mainly aiming at improving the inference cost efficiency, we also measure and compare training times between our method and the baselines, on the PASCAL-Context \cite{chen2014detect}. The results are shown on \hyperref[tab:training]{Table~\ref{tab:training}}. The forward and backward iterations are averaged over 1000 iterations, after 10 warmup iterations, on a single NvidiaV100 GPU, with a batch size of 4.

\begin{table}[!h]
\centering
\caption{Training time comparison of various MTL methods}
\label{tab:training}
\begin{tabular}{@{}lcccc@{}}
\toprule
Method       & Forward (ms) & Backward (ms) & Training time (h) & $\Delta_\text{MTL}$ \\ \midrule
Standard MTL &    60     &   299       &        7.5       &    -4.14       \\ 
MTAN         &    73     &    330      &       8.5        &      -1.78     \\ 
Cross-stitch  &  132    &    454    &         12.3      &      +0.14     \\    
MGDA-UB         &    60     &     568     &         13.2      &   -1.94        \\     
CAGrad       &    60     &    473      &      11.1         &   -2.03        \\ 
PCGrad       &    60    &     495     &       11.6        &    -2.58       \\ 
\midrule
\methodname{}&  76     &   324    &       8.4     &  -1.35    \\ 
\methodname{}&  102    &   376     &       10.1    &  +0.12    \\ 
\methodname{}&  119    &   426     &       11.5     &  +0.42 \\ 
\bottomrule
\end{tabular}
\end{table}

\label{sec:sup}
\end{document}